\def\year{2021}\relax
\documentclass[letterpaper]{article} 
\usepackage{aaai21}  
\usepackage{times}  
\usepackage{helvet} 
\usepackage{courier}  
\usepackage[hyphens]{url}  
\usepackage{graphicx} 
\usepackage{graphicx}  
\usepackage{natbib}  
\usepackage{caption} 
\title{\EECBS{}: A Bounded-Suboptimal Search for Multi-Agent Path Finding}
\author{
Jiaoyang Li,\textsuperscript{\rm 1}\protect\thanks{Jiaoyang Li performed the research during her visit to Monash University.}
Wheeler Ruml,\textsuperscript{\rm 2}
Sven Koenig\textsuperscript{\rm 1}\\
}
\affiliations{
\textsuperscript{\rm 1}University of Southern California, Los Angeles, California, USA\\
\textsuperscript{\rm 2}University of New Hampshire, Durham, New Hampshire, USA\\
jiaoyanl@usc.edu,
ruml@cs.unh.edu,
skoenig@usc.edu
}

\usepackage[switch]{lineno}  %
\usepackage{amsmath}
\usepackage{subfigure}
\usepackage{multirow}
\usepackage{multicol}
\usepackage[linesnumbered,vlined,ruled]{algorithm2e}
\SetArgSty{textnormal} 
\usepackage{comment}
\usepackage{cleveref}

\newcommand{\EECBS}[1]{EECBS}
\newcommand{\OPEN}[1]{OPEN}
\newcommand{\FOCAL}[1]{FOCAL}
\newcommand{\CLEANUP}[1]{CLEANUP}
\newcommand{\citepw}[1]{\citet{#1}}

\begin{document}
\maketitle

\begin{abstract}
Multi-Agent Path Finding (MAPF), i.e., finding collision-free paths for multiple robots, is important for many applications where small runtimes are necessary, including the kind of automated warehouses operated by Amazon. 
CBS is a leading two-level search algorithm for solving MAPF optimally. 
ECBS is a bounded-suboptimal variant of CBS that uses focal search to speed up CBS by sacrificing optimality and instead guaranteeing that the costs of its solutions are within a given factor of optimal. In this paper, we study how to decrease its runtime even further using inadmissible heuristics.  Motivated by
Explicit Estimation Search (EES), we propose Explicit Estimation CBS (\EECBS{}), a new bounded-suboptimal variant of CBS, that uses online learning to obtain inadmissible estimates of the cost of the solution of each high-level node and uses EES to choose which high-level node to expand next. We also investigate recent improvements of CBS and adapt them to \EECBS{}.  We find that \EECBS{} with the improvements runs significantly faster than the state-of-the-art bounded-suboptimal MAPF algorithms ECBS, BCP-7, and eMDD-SAT on a variety of MAPF instances. We hope that the scalability of \EECBS{} enables additional applications for bounded-suboptimal MAPF algorithms. 
\end{abstract}

\section{Introduction}

Multi-Agent Path Finding (MAPF) is the problem of finding collision-free paths for a team of agents in a known environment while minimizing the sum of their travel times. It is inspired by real-world applications such as warehouse logistics~\cite{MaAAMAS17}, airport operations~\cite{LiAIAA19}, UAV traffic management~\cite{HoAAMAS19}, automated valet parking~\cite{OkosoITSC19}, and video games~\cite{LiAAMAS20a}.

CBS~\cite{SharonAIJ15} is a leading two-level search algorithm for solving MAPF optimally.
Its central idea is to plan a path for each agent independently and then resolve collisions between two agents by branching. Each branch is a new candidate plan wherein one agent or the other is forced to find a new path that avoids the chosen collision.
Researchers have made significant progress on speeding up CBS~\cite{BoyarskiICAPS15,BoyarskiIJCAI15,BoyarskiIJCAI20,BoyarskiSoCS20,FelnerICAPS18,GangeICAPS19,LiICAPS19,LiAAAI19a,LiICAPS20,ZhangICAPS20} and designing compilation-based algorithms that use ideas similar to CBS~\cite{SurynekIROS19,LamBHS19,LamICAPS20}.
However, for many applications, 
we need to coordinate hundreds of agents with limited computational resources.  The fact that MAPF is NP-hard to solve optimally~\cite{YuAAAI13} motivates finding bounded-suboptimal solutions in order to reduce the runtime of the search. 

Enhanced CBS (ECBS)~\cite{BarerSoCS14} is a bounded-suboptimal variant of CBS that is guaranteed to find solutions whose costs are no more than a user-specified factor away from optimal. Its bounded suboptimality is achieved by replacing the best-first search in the high and low levels of CBS with focal search~\cite{PearlK82}. Focal search uses an admissible heuristic for bounding the solution cost and another heuristic for determining the distance of nodes to the goal nodes. 
We will demonstrate that ECBS becomes inefficient if these heuristics are negatively correlated.

In this paper, we propose a new bounded-suboptimal variant of CBS, called Explicit Estimation CBS (\EECBS{}).  \EECBS{} replaces focal search with Explicit Estimation Search~\cite{ThayerIJCAI11} on the high level and uses online learning~\cite{ThayerICAPS11} to learn an informed but potentially inadmissible heuristic to guide the high-level search. 
ECBS and \EECBS{} differ from CBS only in the node selection rules used in their high- and low-level searches. Hence, the ideas behind many improvements of CBS, such as bypassing conflicts~\cite{BoyarskiICAPS15}, prioritizing conflicts~\cite{BoyarskiIJCAI15}, symmetry reasoning~\cite{LiAAAI19a,LiICAPS20}, and focusing the high-level search with admissible heuristics~\cite{LiIJCAI19}, might improve ECBS and EECBS as well, and we show how they can be adapted to them.
We empirically evaluate how each improvement affects performance, finding that their combination is best. \EECBS{} with the improvements runs significantly faster than ECBS and BCP-7~\cite{LamICAPS20} and eMDD-SAT~\cite{SurynekSoCS18}, two other state-of-the-art bounded-suboptimal MAPF algorithms.

\section{Preliminaries}

In this section, we formalize the definition of MAPF and introduce the baseline algorithms CBS and ECBS. 

\subsection{Multi-Agent Path Finding (MAPF)}
MAPF has many variants. In this paper, we focus on the variant defined by \citepw{SternSOCS19} that (1) considers vertex and swapping conflicts, (2) uses the ``stay at target'' assumption, and (3) optimizes the sum of costs.
Formally, we define MAPF by a graph 
and a set of $m$ agents \{$a_1, \ldots, a_m\}$. 
Each agent $a_i$ has a start vertex $s_i$ 
and a target vertex $g_i$. 
Time is discretized into timesteps. At each timestep,
every agent can either move to an adjacent vertex or wait at its current
vertex. 
A \emph{path} $p_i$ for agent $a_i$
is a sequence of vertices that are pairwise
adjacent or identical (indicating a wait action),
starting at the start vertex $s_i$ and
ending at the goal vertex $g_i$. 
The \emph{cost} of a path $p_i$ is its length $|p_i|$. 
Agents remain at their goal vertices after they complete their paths. 
We say that two agents have a \emph{conflict}
iff they are at the same vertex or traverse the same edge in opposite directions at the same timestep. 
A \emph{solution} is a set of conflict-free paths $\{p_1, \dots, p_m\}$, one for each agent. 
An \emph{optimal solution} is a solution with the minimum \emph{sum of costs} $\sum_{i=1}^m |p_i|$. 

\subsection{Conflict-Based Search (CBS)}
\subsubsection{Vanilla CBS}
CBS~\cite{SharonAIJ15} is a two-level search algorithm for solving MAPF optimally. 
On the high level, CBS performs a best-first search 
on a binary \emph{constraint tree} (CT). Each CT node $N$ contains a set of
constraints $N.constraints$ that are used to coordinate agents to avoid conflicts and a set of paths $N.paths$, one for each agent, that satisfy the constraints. We use $N.paths[i]$ to denote the path of agent $a_i$.
The \emph{cost} 
of $N$ is the sum of costs of its paths, i.e., $cost(N)=\sum_{i=1}^m|N.paths[i]|$. The root CT node contains an empty set of constraints and a set of shortest paths. 

When expanding a CT node, CBS checks 
for conflicts among its paths. 
If there are none, then the CT node is a goal CT node, and CBS terminates.
Otherwise, CBS chooses one of the conflicts and 
resolves it by \emph{splitting} the CT node into two child CT nodes. 
In each child CT node, one agent from the conflict is prohibited from using the
conflicting vertex or edge at the conflicting timestep by way of an additional constraint. The path of this
agent does not satisfy the new constraint and is replanned with A* on the low level. 
CBS guarantees completeness by eventually exploring both ways of resolving each conflict. It guarantees optimality by performing best-first searches on both its high and low levels.

\subsubsection{Recent Improvements of CBS}

Researchers have proposed many techniques for speeding up CBS while preserving its completeness and optimality.

\emph{Bypassing conflicts}~\cite{BoyarskiICAPS15} is a conflict-resolution technique that, instead of splitting a CT node, modifies the paths of the agents involved in the chosen conflict.
When expanding a CT node $N$ and generating child CT nodes, if the cost of a child CT node $N'$ is equal to $cost(N)$ and the number of conflicts of $N'.paths$ is smaller than the number of conflicts of $N.paths$, then it replaces the paths in $N$ with the paths in $N'$ and discards all generated child CT nodes. 
Otherwise, it splits $N$ as before. It has been shown that bypassing conflicts often produces smaller CTs and decreases the runtime of CBS.

\emph{Prioritizing conflicts}~\cite{BoyarskiIJCAI15} is a conflict-selection technique. 
A conflict is \emph{cardinal}
iff, when CBS uses the conflict to split a CT node $N$, the costs of both child CT nodes are
larger than $cost(N)$.
It is \emph{semi-cardinal} iff the cost of one child CT node is larger than $cost(N)$ and the cost of the other child CT node is equal to $cost(N)$. It is \emph{non-cardinal} iff the costs of both child CT nodes are equal to $cost(N)$. The cardinality of a conflict is determined by building a Multi-Valued Decision Diagram (MDD)~\cite{SharonAIJ13} for each agent (i.e., an acrylic directed graph that consists of all shortest paths of the agent). 
CBS can significantly improve its efficiency by resolving cardinal conflicts first, then semi-cardinal conflicts, and finally non-cardinal conflicts, because generating CT nodes with larger costs first typically improves the lower bound of the CT (i.e., the minimum cost of the CT nodes in the open list) 
faster and thus produce smaller CTs. 

\begin{figure}
    \centering
    \includegraphics[width=0.9\columnwidth]{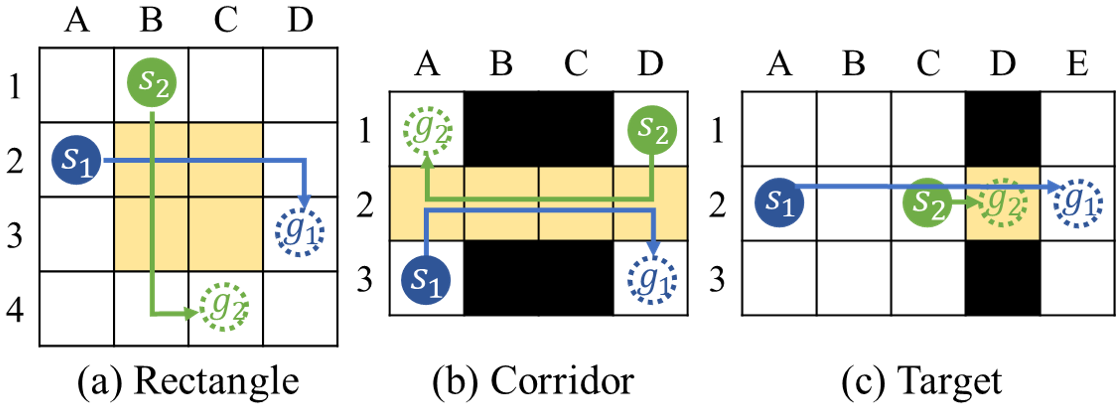}
    \caption{Examples of symmetries on 4-neighbor grids.
    }
    \label{fig:rect}
\end{figure}

\emph{Symmetry reasoning}~\cite{LiAAAI19a,LiICAPS20} is a technique for avoiding to resolve conflicts between the same pair of agents repeatedly due to symmetric paths and conflicts. \Cref{fig:rect}(a) shows an example of rectangle symmetry.  
There exist multiple shortest paths for each agent, each of which can be obtained by changing the order of the individual
\textsc{Right} and \textsc{Down} moves. Any shortest path for
one agent is in conflict with any shortest path for the other
agent. CBS has to try multiple combinations of these paths before realizing that one of the agents has to wait for one timestep. The size of the CT grows exponentially as the size of the yellow rectangular area (i.e., the region where conflicts can occur) increases. Similar behaviors of CBS can be observed in corridor and target symmetries (\Cref{fig:rect}(b) and (c)). Symmetry reasoning identifies each symmetry efficiently and resolves it by a single splitting action with specialized constraints, producing smaller CTs and decreasing the runtime of CBS.

\emph{Weighted Dependency Graph (WDG) heuristic}~\cite{LiIJCAI19} is an admissible heuristic for the high-level search of CBS.
It proceeds by building a weighted dependency graph for each CT node $N$, whose vertices represent agents,
whose edges represent that the two corresponding agents are dependent, i.e., the minimum sum of costs of their conflict-free paths that satisfy $N.constraints$ (which is computed by solving a 2-agent MAPF instance using CBS) is larger than the sum of costs of their paths in $N.paths$ (which are the shortest paths that satisfy $N.constraints$ but are not necessarily conflict-free), and whose edge weights represent the difference between the minimum sum of costs of the conflict-free paths that satisfy $N.constraints$ and the sum of costs of their paths in $N.paths$.
The value of the edge-weighted minimum vertex cover of the graph is then used as an admissible heuristic for the high-level search. Despite the runtime overhead of building the weighted dependency graphs and finding their edge-weighted minimum vertex cover, the addition of the WDG heuristics often produces smaller CTs and decreases the runtime of CBS.

\subsection{Enhanced CBS (ECBS)}

\emph{Focal search} is a bounded-suboptimal search algorithm based on $A^*_\epsilon$~\cite{PearlK82}. 
It maintains two lists of nodes: \OPEN{} and \FOCAL{}. 
\OPEN{} is the regular open list of A*, sorted according to an admissible cost function $f$.
Let $best_{f}$ be the node in
\OPEN{} with the minimum $f$ value and $w$ be a user-specified suboptimality factor.
\FOCAL{} contains those nodes $n$ in \OPEN{} for which $f(n) \leq w \cdot f(best_{f})$, sorted according to a function $d$ that estimates the \emph{distance-to-go}, i.e., the number of hops from node $n$ to a goal node. Focal search always expands a node with the minimum $d$ value in \FOCAL{}.
Since $f(best_{f})$ is a lower bound on the optimal solution cost $C^*$, focal search guarantees that the returned solution cost is at most $w \cdot C^*$.

ECBS~\cite{BarerSoCS14} is a bounded-suboptimal variant of CBS that uses focal search with the same suboptimality factor $w$ on both the high and low levels.
The low level of ECBS finds a bounded-suboptimal path for agent $a_i$ that satisfies the constraints of CT node $N$ and minimizes the number of conflicts with the paths of other agents. 
It achieves this by using a focal search with $f(n)$ being the standard $f(n) = g(n)+h(n)$ of A* and $d(n)$ being the number of conflicts with the paths of other agents. When it finds a solution, it returns not only the path but also the minimum $f$ value $f^i_{\min}(N)$ in
\OPEN{}, indicating a lower bound on the cost of the shortest path for agent $a_i$. For clarity, we denote the cost of the shortest path for agent $a_i$ as $f_{opt}^i(N)$, but this value is in general unknown during search.
Thus,
\begin{equation}
    f^i_{\min}(N) \leq f_{opt}^i(N) \leq |N.paths[i]| \leq w \cdot f^i_{\min}(N). \label{eqn:0}
\end{equation}
Unlike usual bounded-suboptimal searches, the focal search used on the low level of ECBS is to speed up the high-level search, instead of the low-level search itself, as it tries to reduce the number of conflicts that need to be resolved by the high-level search.

The high level of ECBS uses a modified focal search. \OPEN{} is the regular open list of A*, which sorts its CT nodes $N$ according to $lb(N)=\sum_{i=1}^m f^i_{\min}(N)$, indicating a lower bound on the minimum cost of the solutions below CT node $N$. 
From \Cref{eqn:0}, we know that 
    $lb(N) \leq cost(N) \leq w \cdot lb(N)$. 
Let $best_{lb}$ be the node in \OPEN{} with the minimum $lb$ value.
\FOCAL{} contains those CT nodes $N$ in \OPEN{} for which $cost(N) \leq w \cdot lb(best_{lb})$, sorted according to the number of conflicts $h_c(N)$ of $N.paths$, roughly indicating the distance-to-go for the high-level search, i.e.,  the number of splitting actions required to find a solution below CT node $N$. 
Since $lb(best_{lb})$ is a lower bound on the optimal sum of costs, the cost of any CT node in \FOCAL{} is no larger than $w$ times the optimal sum of costs. Thus, once
a solution is found, its sum of costs is also no larger than $w$ times the optimal sum of costs.

\section{Explicit Estimation CBS (EECBS)}

We first analyze the behavior of the high-level focal search of ECBS. We then present our new algorithm \EECBS{}, which uses Explicit Estimation Search on the high level and online learning to estimate the solution cost. 

\begin{figure*}
    \centering
    \includegraphics[height=3cm]{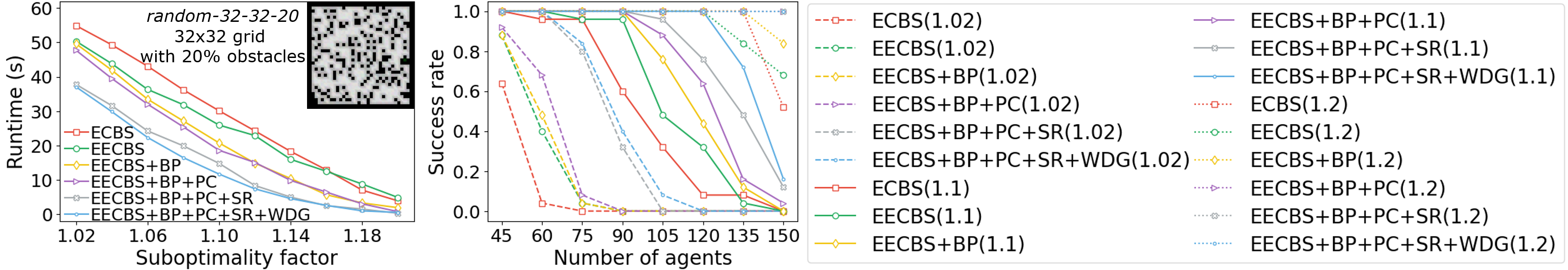}
    \caption{Performance. The left figure plots the average runtime of the algorithms over 200 instances.
    One minute is included in the average for each unsolved instance.
    The right figure plots the \emph{success rate} (i.e., the percentage of the instances solved within the time limit) of the algorithms with suboptimality factors 1.02, 1.10, and 1.20.  
    BP, PC, SR, and WDG are short for bypassing conflicts, prioritizing conflicts, symmetry reasoning, and using the WDG heuristic, respectively.
    }
    \label{fig:success}
\end{figure*}

To evaluate the effectiveness of each technique that we introduce, we test it on 200 standard MAPF benchmarks~\cite{SternSOCS19} with a time limit of one minute per instance. 
In particular, we use map \texttt{random-32-32-20}, a $32\times32$ 4-neighbor 
grid with 20\% randomly blocked cells, shown in \Cref{fig:success}, with the number of agents varying from 45 to 150 in increments of 15.
We use the ``random'' scenarios of the benchmarks, yielding 25 instances for each number of agents.
We vary the suboptimality factor from 1.02 to 1.20 in increments of 0.02.

\subsection{Limitations of ECBS}\label{sec:limitation}

\begin{table}
    \centering
    \resizebox{\columnwidth}{!}{
    \begin{tabular}{c|c|ccccc}
        \multicolumn{2}{c|}{$w$} & 1.04 & 1.08 & 1.12 & 1.16 & 1.20 \\
        \hline
        ECBS & $\Delta lb$ & 0.63 & 0.63 & 0.64 & 0.52 & 0.56 \\ 
        \hline
        \multirow{2}{*}{\EECBS}  & $\Delta lb$ & 5.21 & 2.32 & 1.40 & 0.71 & 0.64 \\ 
        & Cleanup & 39.6\% & 14.4\% & 11.1\% &  0.7\% & 0.0\% \\
    \end{tabular}
    }
    \caption{Lower-bound improvement $\Delta lb$, i.e., the value of $lb(best_{lb})$ when the algorithm terminates minus the $lb$ value of the root CT node.
    Cleanup represents the percentage of expanded CT nodes that are selected from \CLEANUP{}.
    }
    \label{tab:lb}
\end{table}

\begin{figure*}[ht!]
    \centering
    \includegraphics[width=\textwidth]{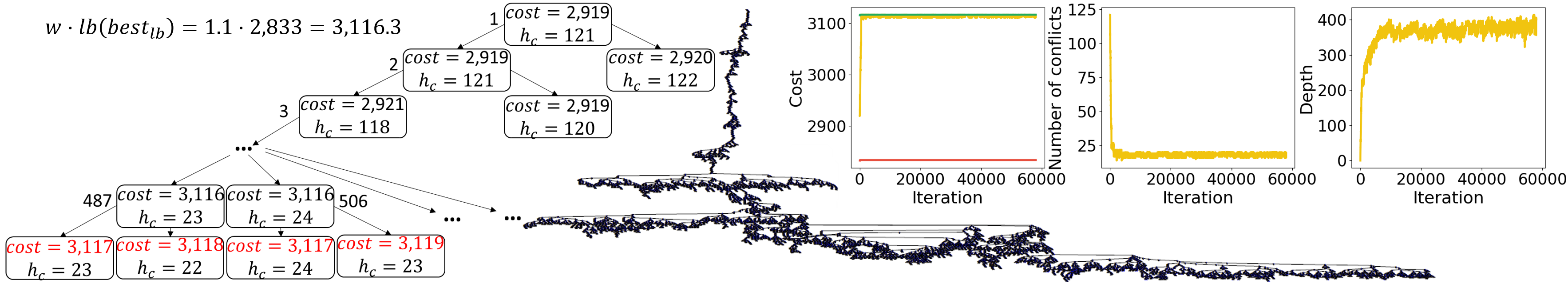}
    \caption{Performance of ECBS on a hard MAPF instance. 
    The left and middle diagrams illustrate the CT tree of ECBS after 506 and 5,000 iterations (i.e., after expanding 506 and 5,000 CT nodes). 
    The right three charts plot the cost, the $h_c$ value, and the depth of the selected CT node at each iteration, respectively, with the red and green lines in the first chart being the lower bound $lb(best_{lb})$ and the suboptimality bound $w \cdot lb(best_{lb})$ at each iteration.
    }\label{fig:ECBS-bad-example}
\end{figure*}

\Cref{fig:ECBS-bad-example} shows the typical behavior of ECBS on a hard MAPF instance, revealing two drawbacks of its high-level focal search. 
The first drawback is that, when selecting CT nodes for expansion, ECBS only considers the distance-to-go and requires the cost of the selected CT node $N$ to be within suboptimality bound $w \cdot lb(best_{lb})$ but ignores the fact that the cost of the solution below CT node $N$ is likely to be larger than $cost(N)$ and thus could also be larger than the suboptimality bound.
In the example of \Cref{fig:ECBS-bad-example}, ECBS first keeps expanding CT nodes roughly along a branch (i.e., CT nodes $1 \rightarrow 2 \rightarrow 3 \rightarrow \dots \rightarrow 487$ in the left diagram), so the cost of the selected CT node keeps increasing and its $h_c$ value keeps decreasing until it expands a CT node $N$ both of whose child CT nodes are not qualified for inclusion in \FOCAL{} (i.e., CT node 487 in the left diagram). As a result, it then expands a neighboring CT node $N'$ whose cost is slightly smaller than $cost(N)$ and whose $h_c$ value is slightly larger than $h_c(N)$. It keeps expanding CT nodes below  CT node $N'$, but, after several iterations, expands another CT node both of whose child CT nodes are not qualified for inclusion in \FOCAL{} (i.e., CT node 506 in the left diagram). 
It repeats numerous times, indicated by the right charts, and as a result, as shown in the middle diagram, ECBS is stuck in a local area of the CT and never gets a chance to explore other parts of the CT. 
This thrashing behavior, in which the negative correlation of $h_c(N)$ and $cost(N)$ causes focal search to repeatedly abandon the children of expanded CT nodes, was noted by \citepw{thayer:ude}. 

The second drawback is that the lower bound of ECBS rarely increases, as shown by the flat red line in the first chart of \Cref{fig:ECBS-bad-example}. This is because, when expanding a CT node $N$, ECBS resolves a conflict and adds new constraints to the generated child CT nodes, so the $lb$ values of the child CT nodes tend to be equal to or larger than $lb(N)$ while the $h_c$ values of the child CT nodes tend to be smaller than $h_c(N)$. As a result,  $best_{lb}$ tends to have a large $h_c$ value. 
This results in $best_{lb}$ being expanded only when \FOCAL{} is almost empty. Since there are many CT nodes of the same cost, ECBS rarely empties \FOCAL{}, and thus its lower bound $lb(best_{lb})$ rarely increases. More statistics can be found in \Cref{tab:lb} (second row). Therefore, if the optimal sum of costs is not within the initial suboptimality bound, ECBS can have difficulty finding a solution within a reasonable time.
While this problematic behavior is similar to that in the first drawback, in that both involve a negative correlation of node values, it is subtly different as it involves the lower bound rather than the cost.

\subsection{Explicit Estimation Search (EES)}

EES~\cite{ThayerIJCAI11} is a bounded-suboptimal search algorithm designed in part to overcome the poor behavior of focal search.
It introduces a third function $\hat{f}$ that estimates, potentially inadmissibly, the cost of the solution below a given node. EES combines estimates of the distance-to-go and the solution cost to predict the expansion of which nodes will lead most quickly to a solution within the given suboptimality factor. If the nodes of interest are not within the current suboptimality bound, it expands the node with the minimum $f$ value to improve the current suboptimality bound.
Formally, EES maintains three lists of nodes: \CLEANUP{}, \OPEN{}, and \FOCAL{}. 
\CLEANUP{} is the regular open list of A*, sorted according to an admissible cost function $f$.
Let $best_{f}$ be the node in \CLEANUP{} with the minimum $f$ value and $w$ be the user-specified suboptimality factor. 
\OPEN{} is also another regular open list of A*, sorted according to a more informed but potentially inadmissible cost function $\hat{f}$.
Let $best_{\hat{f}}$ be the node in \OPEN{} with the minimum $\hat{f}$ value.
\FOCAL{} contains those nodes $n$ in \OPEN{} for which 
$\hat{f}(n) \leq w \cdot \hat{f}(best_{\hat{f}})$, sorted according to a distance-to-go function $d$. 
$\hat{f}(best_{\hat{f}})$ is an estimate of the cost of an optimal solution, so EES suspects that expanding the nodes in \FOCAL{} can lead to solutions that are no more than $w$ times away from optimal.
Let $best_{d}$ be the node in \FOCAL{} with the minimum $d$ value.
When selecting nodes for expansion,
EES first considers $best_{d}$, as expanding nodes with nearby goal nodes should lead to a goal node fast. 
To guarantee bounded suboptimality, EES selects $best_{d}$ for expansion only if $f(best_d) \leq w \cdot f(best_f)$.
If $best_{d}$ is not selected, EES next considers $best_{\hat{f}}$, as it suspects that $best_{\hat{f}}$ lies along a path to an optimal solution.
To guarantee bounded suboptimality, EES selects $best_{\hat{f}}$ for expansion only if $f(best_{\hat{f}}) \leq w \cdot f(best_f)$.
If neither $best_{d}$ nor $best_{\hat{f}}$ is selected, 
EES selects $best_{f}$, which can raise the suboptimality bound $w \cdot f(best_{f})$,
allowing EES to consider $best_{d}$ or $best_{\hat{f}}$ in the next iteration.

\subsection{Explicit Estimation CBS (EECBS)}

We form \EECBS{} by replacing focal search with EES on the high level of ECBS. Formally, the high level of EECBS maintains three lists of CT nodes: \CLEANUP{}, \OPEN{}, and \FOCAL{}. 
\CLEANUP{} is the regular open list of A*, sorted according to the lower bound function $lb$. 
\OPEN{} is another regular open list of A*, sorted according to a potentially inadmissible cost function $\hat{f}$, which estimates the minimum cost of a solutions below a CT node. We use $\hat{f}(N)=cost(N) + \hat{h}(N)$, where $\hat{h}(N)$ is the cost-to-go. We introduce $\hat{h}$ in \Cref{sec:cost-to-go}. 
\FOCAL{} contains those CT nodes $N$ in \OPEN{} for which 
$\hat{f}(N) \leq w \cdot \hat{f}(best_{\hat{f}})$, sorted according to the distance-to go function $h_c$. 
\EECBS{} selects CT nodes from these three lists using the \textsc{selectNode} function: 
\begin{enumerate}
    \item if $cost(best_{h_c}) \leq w \cdot lb(best_{lb})$, then select $best_{h_c}$ (i.e., selected from \FOCAL{});
    \item else if $cost(best_{\hat{f}}) \leq w \cdot lb(best_{lb})$, then select $best_{\hat{f}}$ (i.e., selected from \OPEN{});
    \item else select $best_{lb}$ (i.e., selected from \CLEANUP{}).
\end{enumerate}
Like EES, \EECBS{} selects a CT node $N$ only if its cost is within the current suboptimality bound, i.e., 
\begin{equation}
    cost(N) \leq w \cdot lb(best_{lb}), \label{eqn:2}
\end{equation}%
which guarantees bounded suboptimality.
\EECBS{} uses the same focal search as ECBS on its low level.

\EECBS{} overcomes the first drawback from \Cref{sec:limitation} by taking the potential cost increment below a CT node into consideration in Steps 1 and 2 and selecting a CT node $N$ whose estimated cost $\hat{f}(N)$ is within estimated suboptimality bound $w \cdot \hat{f}(best_{\hat{f}})$. 
It overcomes the second drawback by selecting the CT node with the minimum $lb$ value in Step 3 to raise the lower bound.
\Cref{tab:lb} (third row) shows its empirical behavior in comparison with ECBS. Unlike ECBS, whose lower-bound improvement is always around 0.6, the lower-bound improvement of \EECBS{} increases as the suboptimality factor $w$ decreases. The smaller $w$ is, the less likely a solution is within the initial suboptimality bound and thus the more frequently \EECBS{} selects CT nodes from \CLEANUP{} (as shown in \Cref{tab:lb} (fourth row)).
\Cref{fig:success} shows the runtimes and success rates of ECBS and EECBS. As expected, \EECBS{} (green lines) has a smaller runtime and larger success rate than ECBS (red lines). The improvement increases as $w$ decreases.

\subsection{Online Learning of the Cost-To-Go $\hat{h}$}
\label{sec:cost-to-go}

Our estimate of the minimum cost of the solutions below a given CT node $N$
uses online learning since it does not require preprocessing and allows for instance-specific learning.  \citet{ThayerICAPS11} present a method for learning the cost-to-go during search using the error experienced during node expansions. 
Consider a node $n$ with an admissible cost heuristic $h$ and a distance-to-go heuristic $d$.
The error $\epsilon_d$ of the distance-to-go heuristic $d$, called \emph{one-step distance error}, is defined as $\epsilon_d(n) = d(bc(n)) - (d(n) - 1)$, where $bc(n)$ is the \emph{best child node} of node $n$, i.e., the child node with the smallest $\hat{f}$ value, breaking ties in favor of the child node with the smallest $d$ value.
Similarly, the error $\epsilon_h$ of the cost function $f$, called \emph{one-step cost error}, is defined as $\epsilon_h(n) = h(bc(n)) - (h(n) - c(n, bc(n)))$, where $c(n, bc(n))$ is the cost of moving from node $n$ to node $bc(n)$. These errors can be calculated after every node expansion. 
They use a \emph{global error model} that assumes that the distribution of
one-step errors across the entire search space is uniform
and can be estimated as an average of all observed
one-step errors. Therefore, the search maintains a
running average of the one-step errors observed so far, denoted by $\overline{\epsilon}_d$ and $\overline{\epsilon}_h$. Then, they prove that 
the true cost-to-go of node $n$ can be approximated by
$\hat{h}(n) = h(n) + \frac{d(n)}{1 - \overline{\epsilon}_d(n)} \cdot \overline{\epsilon}_h(n)$ \cite{ThayerICAPS11}.

We apply this method to \EECBS{}.
Since \EECBS{} does not have an admissible cost heuristic $h$, 
we define $\epsilon_d$ of CT node $N$ as $\epsilon_d(N) = h_c(bc(N))- (h_c(N) - 1)$ and $\epsilon_h$ of CT node $N$ as $\epsilon_h(N) = cost(bc(N)) - cost(N)$. Then, 
$\hat{h}(N) = \frac{h_c(N)}{1 - \overline{\epsilon}_d(N)} \cdot \overline{\epsilon}_h(N)$. 
Thus $\hat{h}(N)$ is linear in $h_c(N)$, indicating that the larger the number of conflicts of a CT node, the higher the potential cost increments below the CT node could be.

\Cref{alg} shows the pseudo-code of the high-level search of \EECBS{}. For now, we ignore Lines~\ref{line:root-h}, \ref{line:h1}-\ref{line:SR}, and \ref{line:bypass1}-\ref{line:bypass2} since they will be introduced in the next section. Compared to the high-level search of ECBS, \EECBS{} changes the \textsc{pushNode} and \textsc{selectNode} functions and adds an \textsc{updateOneStepErrors} function at the end of each iteration to update the one-step errors (Line~\ref{line:error}).

\begin{algorithm}[t]
\SetKw{Continue}{continue}
\SetKw{Break}{break}
\caption{High-level search of \EECBS{}.}\label{alg}
\small
Generate root CT node $R$\;
\textsc{computeWDGHeuristic}($R$)\; \label{line:root-h}
$\textsc{pushNode}(R)$\;
\While{\OPEN{} is not empty}{
  $P \leftarrow$ \textsc{selectNode}()\; \label{line:selectNode}
  \If{$P.conflicts$ is empty \label{line:goal-test}}{
    \KwRet $P$.paths\tcp*{P is a goal node}
  }
  \If{$P$ is selected from \CLEANUP{} 
    \textnormal{and} the WDG heuristic of $P$ has not been computed yet \label{line:h1}}{
    \textsc{computeWDGHeuristic}($P$)\;
    \textsc{pushNode}($P$)\;
    \Continue\;\label{line:h2}
  }
  \textsc{conflictPrioritization}($P.conflicts$)\; \label{line:PC}
  \textsc{symmetryReasoning}($P.conflicts$)\; \label{line:SR}
  $conflict \leftarrow \textsc{ChooseConflict}(P.conflicts)$\;
  $constraints \leftarrow \textsc{resolveConflict}(conflict)$\;
  $children \leftarrow \emptyset$\;
  \For{$constraint$ in $constraints$}{
    $Q \leftarrow \textsc{generateChild}(P, constraint)$\;
    
    \If(\tcp*[f]{Bypassing}){$P$ is not selected from \CLEANUP{} 
    \textnormal{and} $\forall i \; |Q.paths[i]| \leq w \cdot f^i_{\min}(P)$ 
    \textnormal{and} $cost(Q) \leq w \cdot lb(best_{lb})$ 
    \textnormal{and} $h_c(Q) < h_c(P)$ \label{line:bypass1}} {
      $P.paths \leftarrow Q.paths$\;
      $P.conflicts \leftarrow Q.conflicts$\;
      Go to Line~\ref{line:goal-test}\;\label{line:bypass2}
    }
    Add $Q$ to $children$\;
  }
  \For{$Q$ in $children$}{
    {$\textsc{pushNode}(Q)$\;}
  }
  \textsc{updateOneStepErrors}($P, children$)\; \label{line:error}
}
\KwRet ``No solutions''\;
\end{algorithm}

\section{Bringing CBS Improvements to \EECBS{}}

We now show how we can incorporate recent CBS improvements into \EECBS{}. We introduce these techniques one by one and, for each one, evaluate its effectiveness by adding it to the best version of \EECBS{} so far and showing the resulting performance in \Cref{fig:success}. 

\subsection{Bypassing Conflicts}

In CBS, the paths of every CT node $N$ are the shortest paths that satisfy $N.constraints$. However, in \EECBS{}, the paths can be bounded-suboptimal. 
Therefore, we can resolve more conflicts with the bypassing conflicts technique in \EECBS{} if we relax the conditions of accepting bypasses. In addition, since bypassing conflicts resolves conflicts without adding any constraints, it does not change the $lb$ value of a CT node. Therefore, it may not be helpful if the purpose of expanding a CT node $N$ is to improve the lower bound, i.e., CT node $N$ is selected from \CLEANUP{}.
Formally, when expanding a CT node $N$ and generating its child CT nodes, \EECBS{} replaces the paths of $N$ with the paths of a child CT node $N'$ and discards all generated child CT nodes iff 
(1) CT node $N$ is not selected from \CLEANUP{},
(2) the cost of every path of CT node $N'$ is within the suboptimality bound of the corresponding agent in CT node $N$, i.e., $\forall i \; |N'.paths[i]| \leq w \cdot f^i_{\min}(N)$, (3) the cost of CT node $N'$ is within suboptimality bound $w \cdot  lb(best_{lb})$,
and (4) the number of conflicts decreases, i.e.,  $h_c(N') < h_c(N)$.
The first condition avoids wasting time in applying the bypassing conflicts technique to CT nodes that are selected from \CLEANUP{}.  
The second and third conditions ensure that \Cref{eqn:0,eqn:2} hold after replacing the paths, which in turn guarantees bounded suboptimality.
The last condition avoids deadlocks, as in the standard CBS bypassing technique.
See Lines~\ref{line:bypass1}-\ref{line:bypass2} of \Cref{alg}.\footnote{There are many other ways to relax the conditions. For example, one can omit the third condition and re-insert CT node $N$ into the appropriate lists (instead of keeping expanding it) after applying the bypassing conflicts technique. We leave it for future work to evaluate different ways of relaxing the conditions of accepting bypasses for \EECBS{}.}   

\begin{table}
    \centering
    \resizebox{\columnwidth}{!}{
    \begin{tabular}{c|ccccc} 
        $w$ & 1.04 & 1.08 & 1.12 & 1.16 & 1.20 \\
        \hline
        CBS bypassing & 0.086 & 0.107 & 0.110 & 0.116 & 0.108 \\
        Relaxed bypassing & 0.091 & 0.114 & 0.104 & 0.131 & 0.126 
    \end{tabular}}
    \caption{Average number of accepted bypasses per 
    CT node.
    }
    \label{tab:bypass}
\end{table}

Empirically, we compare the effectiveness of \EECBS{} with relaxed bypassing and \EECBS{} with CBS bypassing.
\Cref{tab:bypass} reports the average number of accepted bypasses per expanded CT node. Relaxed bypassing accepts more bypasses than CBS bypassing, and the difference increases as the suboptimality factor $w$ increases.
The yellow and green lines in \Cref{fig:success} show the performance of \EECBS{} with and without relaxed bypassing. Relaxed bypassing improves the performance of \EECBS{} for all values of $w$ and all numbers of agents. In general, the improvement increases with $w$.

\subsection{Prioritizing Conflicts}

In CBS, the prioritizing conflicts technique (PC) tries to improve the costs of CT nodes faster since the cost of a CT node also serves as a lower bound on the minimum cost of the solutions below this CT node. In \EECBS{}, however, the cost of a CT node is different from its lower bound. Therefore, we reformulate cardinal, semi-cardinal, and non-cardinal conflicts.  
A conflict is \emph{cardinal}
iff, when CBS uses the conflict to split CT node $N$ and generates two child CT nodes $N'$ and $N''$, both $\sum_{i=1}^mf_{opt}^i(N')$ and $\sum_{i=1}^mf_{opt}^i(N'')$ are
larger than $\sum_{i=1}^mf_{opt}^i(N)$.
The changes to the definitions of semi-cardinal and non-cardinal conflicts are similar.
Like CBS, \EECBS{} performs PC before it chooses conflicts (Line~\ref{line:PC}) and uses MDDs to classify conflicts.
Since the construction of MDDs induces runtime overhead and not all cardinal conflicts are important to \EECBS{}, 
\EECBS{} classifies a conflict between agents $a_i$ and $a_j$ only when (1) the CT node $N$ is selected from \CLEANUP{} or (2) at least one of the path costs is equal to its lower bound, i.e., $|N.paths[i]|=f_{\min}^i(N)$ or $|N.paths[j]|=f_{\min}^j(N)$. 
PC is applied in Case (1) because resolving cardinal conflicts tends to raise the lower bound in this case. It is applied in Case (2) because resolving cardinal conflicts can increase the agents' path costs in this case (which makes the costs of the resulting child CT nodes closer to the sum of costs of the solutions below them). \EECBS{} selects cardinal conflicts first, then semi-cardinal conflicts, then non-cardinal conflicts, and finally unclassified conflicts.
By comparing the purple and yellow lines in \Cref{fig:success}, we see that PC improves the performance of \EECBS{}.

\subsection{Symmetry Reasoning}

We adapt symmetry reasoning techniques to \EECBS{} with almost no changes. \EECBS{} performs symmetry reasoning before it chooses conflicts (Line~\ref{line:SR}).
Rectangle symmetry occurs only if the paths of both agents are the shortest because, otherwise, one of the agents can simply perform a wait action to avoid the rectangle symmetry. Therefore, for a given conflict between agents $a_i$ and $a_j$ at CT node $N$, we apply rectangle symmetry reasoning only when the paths of both agents are provably the shortest, i.e., $|N.paths[i]| = f_{\min}^i(N)$ and $|N.paths[j]| = f_{\min}^j(N)$. 
Corridor and target symmetries can occur even if the paths of the agents are suboptimal. Therefore, we apply corridor and target symmetry reasoning to all conflicts. 
By comparing the grey and purple lines in \Cref{fig:success}, we see that the symmetry reasoning technique significantly improves the performance of \EECBS{}.

\subsection{WDG Heuristic}

\begin{table}
    \centering
    \resizebox{\columnwidth}{!}{
    \begin{tabular}{c|c|ccccc} 
        \multicolumn{2}{c|}{$w$} & 1.04 & 1.08 & 1.12 & 1.16 & 1.20 \\
        \hline
        \multirow{2}{*}{No WDG}  & $\Delta lb$ & 36.1 & 23.4 & 12.9 & 7.4 & 3.9 \\ 
        & Cleanup & 84.0\% & 74.0\% & 56.6\% &  40.4\% & 11.6\% \\
        \hline
        \multirow{2}{*}{CBS WDG}  & $\Delta lb$ & 54.5 & 45.0 & 37.0 & 30.1 & 25.8 \\ 
        & Cleanup & 49.9\% & 46.5\% & 25.1\% &  17.8\% & 5.7\% \\
        & WDG time & 87.9\% & 84.5\% & 80.1\% &  76.4\% & 65.3\% \\
        \hline
        \multirow{3}{*}{Adaptive WDG}  & $\Delta lb$ & 54.8 & 45.2 & 36.8 & 29.1 & 25.3 \\ 
        & Cleanup & 53.2\% & 44.0\% & 27.2\% &  14.4\% & 8.7\% \\
        & WDG time & 77.6\% & 62.6\% & 56.9\% &  43.2\% & 24.1\% 
    \end{tabular}
    }
    \caption{\emph{Lower-bound improvement} $\Delta lb$, i.e., the $lb + h$ value of the best CT node in \CLEANUP{} when \EECBS{} terminates minus the $lb$ value of the root CT node.
    WDG time is the percentage of runtime spent on computing WDG heuristics.
    }
    \label{tab:wdg}
\end{table}

Since a path of a CT node $N$ in \EECBS{} is not necessarily the shortest one and, for each agent $a_i$, $f_{\min}^i(N)$ could be smaller than $f_{opt}^i(N)$, we need to modify the WDG heuristic as follows.   
When \EECBS{} computes the WDG heuristic for a CT node $N$, it builds a weighted dependency graph $G=(V, E)$. Each vertex $i \in V$ corresponds to an agent $a_i$.
The edge weight on each edge $(i,j) \in E$ is equal to the minimum sum of costs of the conflict-free paths for agents $a_i$ and $a_j$ that satisfy $N.constraints$ minus $f_{opt}^i(N)+f_{opt}^j(N)$ (instead of $|N.paths[i]|+|N.paths[j]|$). 
Like CBS, we compute each edge weight by solving a 2-agent MAPF instance (with the constraints in $N.constraints$) using CBS: $f_{opt}^i(N)$ and $f_{opt}^j(N)$ are equal to the costs of the paths of agents $a_i$ and $a_j$ of the root CT node of CBS, and the minimum sum of costs of the conflict-free paths for agents $a_i$ and $a_j$ is equal to the cost of the solution returned by CBS.
Since computing the edge weights for all pairs of agents is time-consuming, we follow CBS by computing the weight for an edge $(i, j)$ only when the paths $N.paths[i]$ and $N.paths[j]$ are in conflict (as the weight is more likely to be larger than 0 in this case). We delete the other edges and the vertices that have no edges. 
Let $h_{\mathrm{WDG}}(N)$ be the value of the edge-weighted minimum vertex cover of $G$~\cite{LiIJCAI19}.
By the proof in \cite{LiIJCAI19}, we know that $\sum_{i=1}^m f_{opt}^i(N) + h_{\mathrm{WDG}}(N)$ is a lower bound on the minimum sum of costs of the solutions below CT node $N$. 
Since 
\begin{align*}
        & \; \textstyle \sum_{i=1}^m f_{opt}^i(N) + h_{\mathrm{WDG}}(N) \\
=       &  \; \textstyle lb(N) + \sum_{i=1}^m (f_{opt}^i(N) - f_{\min}^i(N)) + h_{\mathrm{WDG}}(N) \\
\geq    &  \; \textstyle lb(N) + \sum_{i \in V} (f_{opt}^i(N) - f_{\min}^i(N)) + h_{\mathrm{WDG}}(N), 
\end{align*}
$h(N) = \sum_{i \in V} (f_{opt}^i(N) - f_{\min}^i(N)) + h_{\mathrm{WDG}}(N)$ is admissible, and we can thus use $lb(N)+h(N)$ to sort the CT nodes in \CLEANUP{} and compute the lower bound.

Since computing the WDG heuristic for a CT node is time-consuming, \citepw{LiIJCAI19} suggest to calculate the heuristic lazily by computing a cheaper but less informed heuristic (e.g., using pathmax) when generating a CT node $N$. Only when CT node $N$ is selected for expansion is the WDG heuristic computed and $N$ re-inserted into the appropriate lists. Here, we follow the same scheme, as shown on Lines~\ref{line:h1}-\ref{line:h2}. But, instead of computing the WDG heuristic for all CT nodes, we only compute it for CT nodes selected from \CLEANUP{}, as the purpose of computing the WDG heuristic is to improve the lower bound. In addition, we also compute the WDG heuristic for the root CT node (Line~\ref{line:root-h}), as this can provide a higher lower bound to begin with.

\Cref{tab:wdg} reports the lower-bound improvements of \EECBS{} without the WDG heuristic technique (denoted as No WDG), with the WDG heuristic being computed for all CT nodes (denoted as CBS WDG), and with the WDG heuristic technique introduced above (denoted as Adaptive WDG). 
Compared to No WDG, Adaptive WDG selects CT nodes less frequently from \CLEANUP{} but obtains a larger lower-bound improvement.
Compared to CBS WDG, Adaptive WDG obtains similar lower-bound improvements but spends less time on computing the WDG heuristic.
By comparing the blue and grey lines in \Cref{fig:success}, we see that adaptive WDG improves the performance of \EECBS{} for small and moderately large suboptimality factors.

\begin{figure}[t!]
    \centering
    \includegraphics[width=0.6\columnwidth]{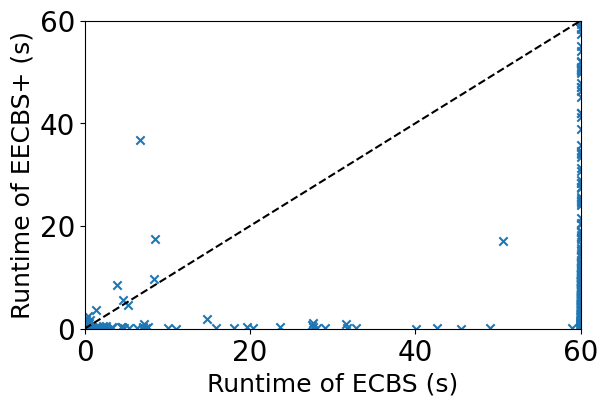}
    \caption{
    Runtimes of ECBS and \EECBS{}+ on the random map with $m$ varying from 45 to 150 and $w$ varying from 1.02 to 1.20. The runtime of each unsolved instance is set to 60 seconds.
    Among the 2,000 instances, 475 instances are solved by neither algorithm; 444 instances are solved only by \EECBS{}+; and 0 instances are solved only by ECBS. 
    }\label{fig:runtimes}
\end{figure}
\begin{figure*}[ht!]
    \centering
    \includegraphics[width=\textwidth]{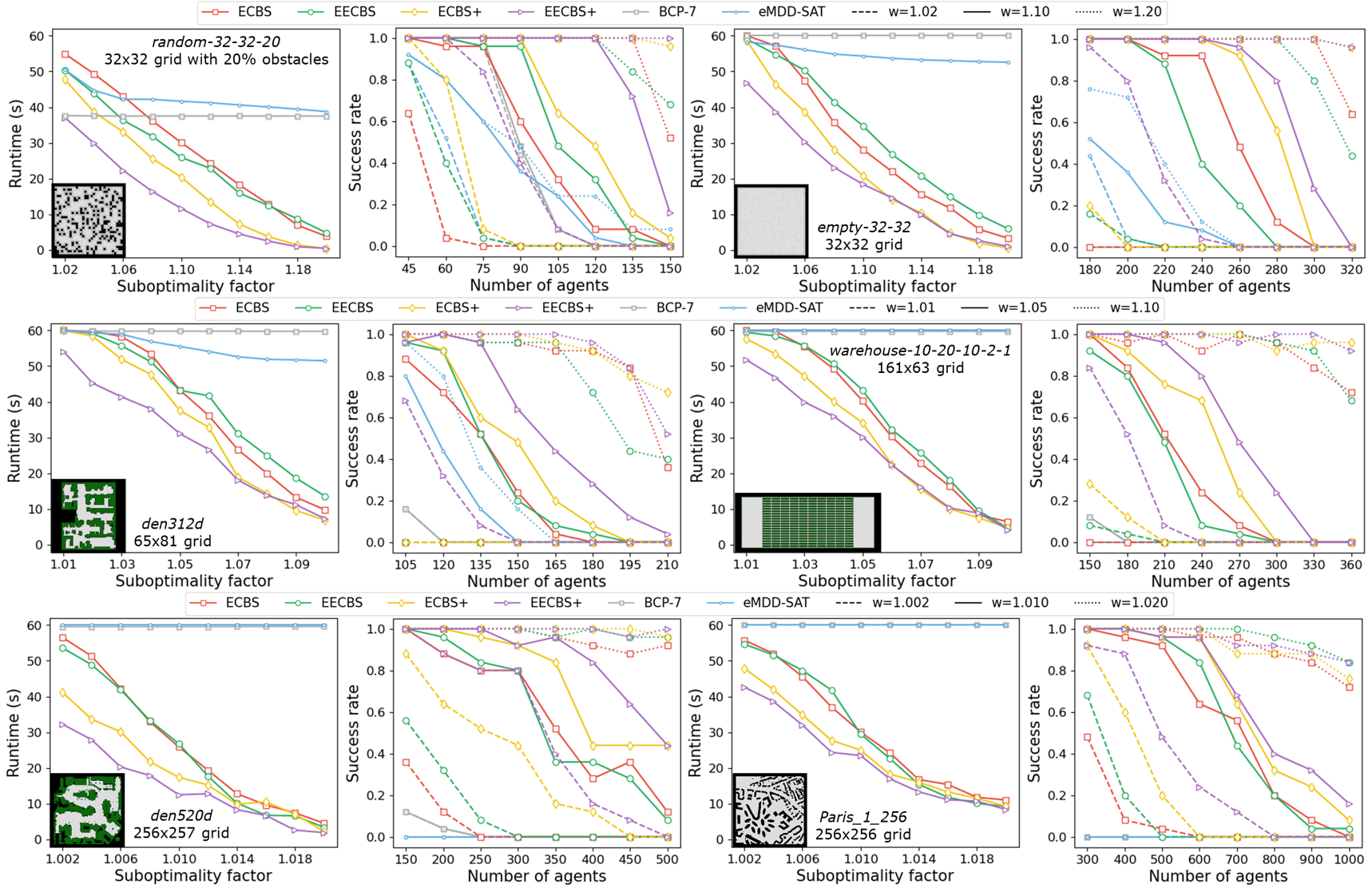}
    \caption{Performance of ECBS, \EECBS{}, ECBS+ (ECBS with all improvements), \EECBS{}+ (\EECBS{} with all improvements), BCP-7, and eMDD-SAT. All results are presented in the same format as in~\protect\Cref{fig:success}. The legend for each row is shown above it.
    The algorithms are indicated by the legend with the same color and the same marker style, 
    while the suboptimality factors of the lines in the success rate figures are indicated by the legend with the same line style.
    Since some algorithms solve (almost) zero instances, 
    their lines overlap at the top of the runtime figures and the bottom of the success rate figures:
    the grey lines are hidden by the blue lines in the runtime figures of maps \texttt{warehouse-10-2-10-2-1}, \texttt{den520d}, and \texttt{Pairs\_1\_256}; 
    the red and green dashed lines are hidden by the yellow dashed line in the success rate figure of map \texttt{den312d}; and many of the grey/ blue dashed/solid/dotted lines are hidden by other lines at the bottom of many of the success rate figures.
    }\label{fig:all}
\end{figure*}
\section{Empirical Evaluation}
\label{sec:exp}

We evaluate the algorithms on 6 maps of different sizes and structures from the MAPF benchmark suite~\cite{SternSOCS19} with 8 different numbers of agents per map.
We use the ``random'' scenarios,
yielding 25 instances for each map and number of agents. 
We evaluate 10 different values of $w$ for each setting, and the values of $w$ decrease as the map size increases.
The algorithms are implemented in C++, and the experiments are conducted on Ubuntu 20.04 LTS on an Intel Xeon 8260 CPU with a memory limit of 16 GB and a time limit of 1 minute. The code is available at \url{https://github.com/Jiaoyang-Li/EECBS}.

As shown in \Cref{fig:all}, \EECBS{} (green lines) outperforms ECBS (red lines) on some maps but shows a similar or even slightly worse performance on other maps. 
The four improvements improve the performance of both ECBS and \EECBS{}, but \EECBS{} benefits more from them. 
As a result, \EECBS{}+ (\EECBS{} with all improvements, purple lines) significantly outperforms the other algorithms on all six maps in terms of both runtime and success rate. When comparing the dashed lines in the success rate figures, we observe that, in many cases (e.g., for 60 agents on the random map and 180 agents on the empty map), \EECBS{}+ is able to solve almost all instances within the runtime limit 
while ECBS solves only a few or even no instances. For a given success rate, \EECBS{}+ is able to solve instances with up to twice the number of agents than ECBS (e.g., on map \texttt{den520d}). 

We provide more details of the experiment on the random map in \Cref{fig:runtimes}. \EECBS{}+ runs faster than ECBS in most cases and never fails to solve an instance that is solved by ECBS.
In addition,  when comparing the solution quality, the average solution costs of ECBS and \EECBS{}+ over the 1,081 instances solved by both algorithms are 1,967 and 1,958, respectively, indicating that the improvements used in \EECBS{}+ do not sacrifice the solution quality.

We omit comparisons with the many search-based bounded-suboptimal MAPF algorithms that have already been shown to perform worse than ECBS~\cite{AljalaudSoCS13,BarerSoCS14,WagnerPhD15}.
However, there are two recent reduction-based bounded-suboptimal MAPF algorithms, namely BCP-7~\cite{LamICAPS20} and eMDD-SAT~\cite{SurynekSoCS18}. BCP-7 can outperform CBS when finding optimal solutions~\cite{LamICAPS20}, and eMDD-SAT can outperform ECBS in some domains~\cite{SurynekSoCS18}. We therefore compare them with our algorithms. We modify the search of the ILP solver used by BCP-7 from a best-first search (which is more beneficial for finding optimal solutions) to a depth-first search with restarts (which is more beneficial for finding suboptimal solutions). As shown in \Cref{fig:all}, BCP-7 and eMDD-SAT outperform ECBS on the two small maps when the suboptimality factor is small. But for larger suboptimality factors or larger maps, they perform worse than ECBS. \EECBS{}+ performs better than them on all six maps. 

\section{Conclusion}

We proposed a new bounded-suboptimal MAPF algorithm \EECBS{} that uses online learning to estimate the cost of the solution below each high-level node and uses EES to select high-level nodes for expansion. We further improve it by adding bypassing conflicts, prioritizing conflicts, symmetry reasoning, and using the WDG heuristic. With these improvements, \EECBS{} significantly outperforms the state-of-the-art
bounded-suboptimal MAPF algorithms ECBS, BCP-7, and eMDD-SAT. 
Within one minute, it is able to find solutions 
that are provably at most 2\% worse than optimal for large MAPF instances with up to 1,000 agents, while, on the same map, state-of-the-art optimal algorithms can handle at most 200 agents~\cite{LamICAPS20}. 
We hope that the scalability of \EECBS{} enables additional applications for bounded-suboptimal MAPF algorithms.


\section*{Acknowledgments}
The research at the University of Southern California was supported by the National Science Foundation (NSF) under grant numbers 1409987, 1724392, 1817189, 1837779, and 1935712 as well as a gift from Amazon. 
The research at the University of New Hampshire was supported by NSF-BSF grant 2008594. 
The views and conclusions contained in this document are those of the authors and should not be interpreted as representing the official policies, either expressed or implied, of the sponsoring organizations, agencies, or the U.S. government.

\bibliography{references}
\end{document}